\documentclass[runningheads]{llncs}
\usepackage{graphicx}
\usepackage{amssymb}
\usepackage{amsmath}
\usepackage{mathtools}
\usepackage[utf8]{inputenc}
\usepackage{amsmath}
\usepackage{xcolor}
\DeclareMathOperator{\vect}{vec}
\begin{document}
\newcommand\ie {{\it i.e., }}
\newcommand\eg {{\it e.g., }}
\newcommand\etc{{\it etc.}}
\newcommand\cf {{\it cf. }}
\newcommand\etal {{\it et al.}}
\newcommand\eq {{\it Eq.}}
\newcommand{\blue}[1]{\textcolor[RGB]{0, 0, 0}{#1}}

\def\rvx{{\mathbf{x}}}
\def\rvy{{\mathbf{y}}}
\def\rvz{{\mathbf{z}}}
\def\rvc{{\mathbf{c}}}
\def\rvh{{\mathbf{\h}}}
\def\rvv{{\mathbf{\beta}}}
\def\rvw{{\mathbf{w}}}
\DeclarePairedDelimiterX{\inp}[2]{\langle}{\rangle}{#1, #2}
\title{Using Causal Analysis for Conceptual Deep Learning Explanation}

%\author{First Author\inst{1}\orcidID{0000-1111-2222-3333} \and
%Second Author\inst{2,3}\orcidID{1111-2222-3333-4444} \and
%Third Author\inst{3}\orcidID{2222--3333-4444-5555}}
%

%% Non anon authors
\author{Sumedha Singla \inst{1} \and
Stephen Wallace \inst{2} \and
Sofia Triantafillou \inst{3} \and \\
Kayhan Batmanghelich \inst{3}
}
%index{Jena, Rohit}
%index{Singla, Sumedha}
%index{Batmanghelich, Kayhan}

\institute{
Computer Science Department, University of Pittsburgh, USA
\and
University of Pittsburgh School of Medicine, University of Pittsburgh, USA
\and Department of Biomedical Informatics, University of Pittsburgh, USA
}
\authorrunning{Singla \etal}


%\institute{ }
%
\maketitle              % typeset the header of the contribution
\begin{abstract}
Model explainability is essential for the creation of trustworthy Machine Learning models in healthcare. An ideal explanation resembles the decision-making process of a domain expert and is expressed using concepts or terminology that is meaningful to the clinicians.
To provide such explanation, we first associate the hidden units of the classifier to clinically relevant concepts. We take advantage of radiology reports accompanying the chest X-ray images to define concepts. We discover sparse associations between concepts and hidden units using a linear sparse logistic regression. To ensure that the identified units truly influence the classifier's outcome, we adopt tools from Causal Inference literature and, more specifically, mediation analysis through counterfactual interventions. Finally, we construct a low-depth decision tree to translate all the discovered concepts into a straightforward decision rule, expressed to the radiologist. We evaluated our approach on a large chest x-ray dataset, where our model produces a global explanation consistent with clinical knowledge.  
\end{abstract}

\section{Introduction}

% first paragraph
Machine Learning, specifically, Deep Learning (DL) methods are increasingly adopted in healthcare applications. Model explainability is essential to build trust in the AI system~\cite{glass2008toward} and to receive clinicians' feedback. Standard explanation methods for image classification delineates regions in the input image that significantly contribute to the model's outcome~\cite{lundberg2017unified,ribeiro2016should,selvaraju2017grad}.  However, it is challenging to explain \emph{how} and \emph{why} variations in identified regions are relevant to the model's decision. Ideally, an explanation should resemble the decision-making process of a domain expert. This paper aims to map a DL model's neuron activation patterns to the radiographic features and constructs a simple rule-based model that partially explains the Black-box.

%second 

Methods based on feature attribution have been commonly used for explaining DL models for medical imaging~\cite{basu2020deep}. However, an alignment between feature attribution and radiology concepts is difficult to achieve, especially when a single region may correspond to several radiographic concepts. Recently, researchers have focused on providing explanations in the form of human-defined concepts~\cite{bau2017network,kim2018interpretability,zhou2018interpretable}. In medical imaging, such methods have been adopted to derive an explanation for breast mammograms ~\cite{yeche2019ubs}, breast histopathology~\cite{graziani2020concept} and cardiac MRIs~\cite{clough2019global}. A major drawback of the current approach is their dependence on explicit concept-annotations, either in the form of a representative set of images~\cite{kim2018interpretability} or semantic segmentation~\cite{bau2017network}, to learn explanations. Such annotations are expensive to acquire, especially in the medical domain. 
We use weak annotations from radiology reports to derive concept annotations. Furthermore, these methods measure correlations between concept perturbations and classification predictions to quantify the concept's relevance. However, the neural network may not use the discovered concepts to arrive at its decision.  We borrow tools from causal analysis literature to address that drawback~\cite{NEURIPS2020_92650b2e}.  

% third paragraph
In this work, we used radiographic features mentioned in radiology reports to define concepts. Using a National Language Processing (NLP) pipeline, we extract weak annotations from text and classify them based on their positive or negative mention~\cite{irvin2019chexpert}. Next, we use sparse logistic regression to identify sets of hidden-units correlated with the presence of a concept. To quantify the causal influence of the discovered concept-units on the model's outcome, we view concept-units as a \emph{mediator} in the treatment-mediator-outcome framework~\cite{imai2011commentary}. Using measures from mediation analysis, we provide an effective ranking of the concepts based on their causal relevance to the model's outcome. Finally, we construct a low-depth decision tree to express discovered concepts in simple decision rules, providing the global explanation for the model. The rule-based nature of the decision tree resembles many decision-making procedures by clinicians.

%Finally, we use a decision tree funcfor a partial global explanation of the model. We adopt the decision tree approach as they are easy to explain, and their rule-based nature resembles many decision-making procedures by clinicians. 

\section{Method}
We consider a pre-trained \emph{black-box} classifier $f: \rvx \rightarrow \rvy$ that takes an image $\rvx$ as input and process it using a sequence of hidden layers to produce a final output $\rvy \in \mathbb{R}^D$. Without loss of generality, we decompose function $f$ as $\Phi_2 \circ \Phi_1(\rvx)$, where $\Phi_1(\rvx) \in \mathbb{R}^L$ is the output of the initial few layers of the network and $\Phi_{2}$ denotes the rest of the network. 
We assume access to a dataset $\mathcal{X} = \{(\rvx_n, \rvy_n, \rvc_n)\}^N$, where $\rvx_n$ is input image, $\rvy_n$ is a $d$-dimensional one-hot encoding of the class labels and $\rvc_n \in \mathbb{R}^K$ is a $k$-dimensional concept-label vector. We define concepts as the radiographic observations mentioned in radiology reports to describe and provide reasoning for a diagnosis. We used a NLP pipeline~\cite{irvin2019chexpert} to extract concept annotations. The NLP pipeline follows a rule-based approach to extract and classify observations from the free-text radiology report. The extracted $k^{th}$ concept-label $\rvc_n [k]$ is either 0 (negative-mention), 1(positive-mention) or -1 (uncertain or missing-mention).  
 %To achieve this, we learn a \emph{surrogate} function $g(\cdot)$ that learns a mapping from the concepts to the class-labels. 
 An overview of our method is shown in Fig.~\ref{fig:model}. Our  method consists of three sequential steps: 
 
  \begin{figure}[htbp]
        \centerline{
        \includegraphics[width=0.89\textwidth]{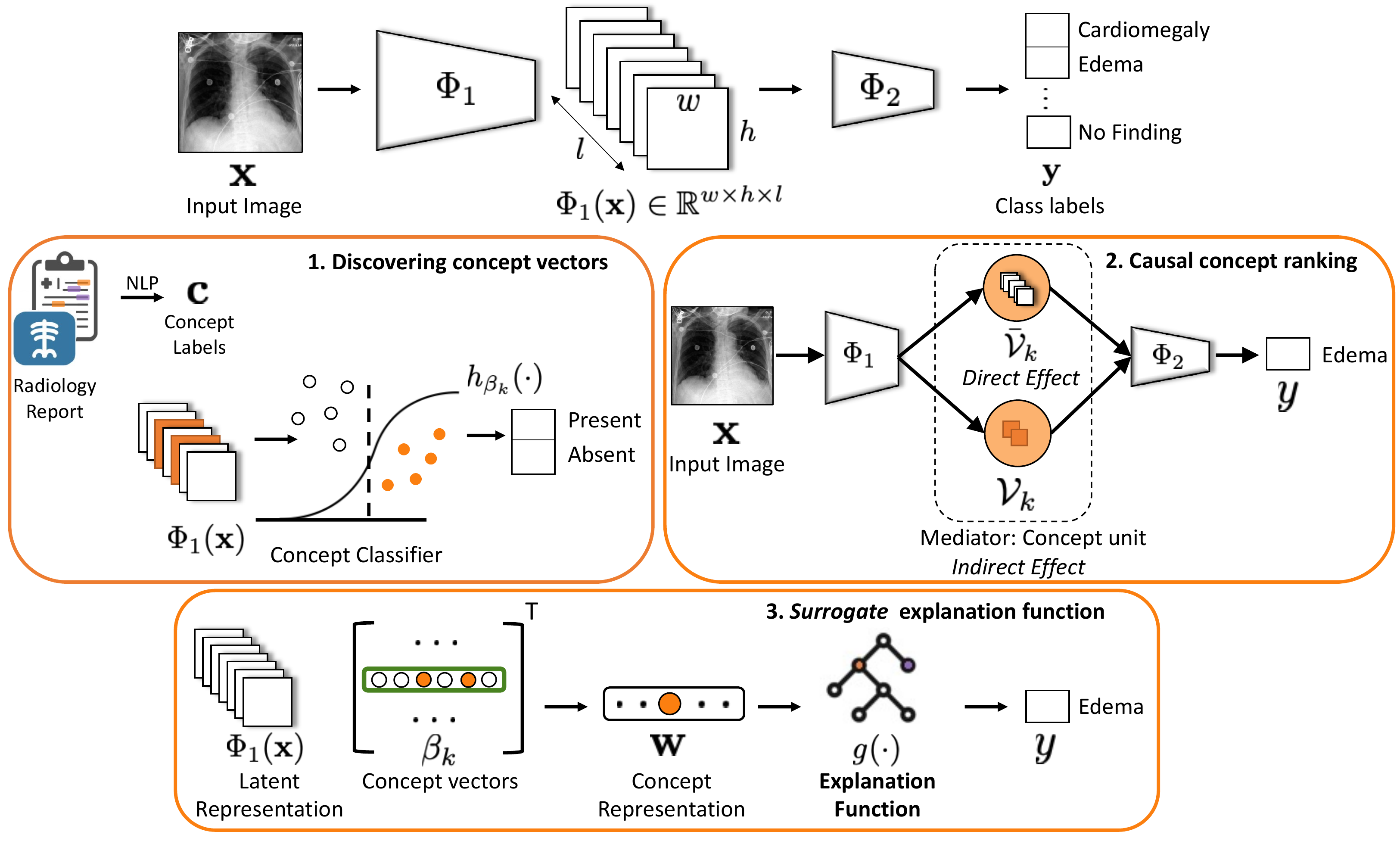}
        }
        \caption{Method overview: We provide explanation for the black-box function $f(\rvx)$ in-terms of concepts, that are radiographic observations mentioned in radiology reports. 1) The intermediate representation  $\Phi_1(\rvx)$ is used to learn a sparse logistic regression $h_{\rvv_k}(\cdot)$ to classify $k^{th}$ concept. 2) The non-zero coefficients of $\rvv_k$ represents a set of concept units $\mathcal{V}_k$ that serves as a mediator in the causal path connecting input $\rvx$ and outcome $y$. 3) A decision tree function is learned to map concepts to class labels.}
        \label{fig:model}
    \end{figure}

(1) \emph{Concept associations}: We seek to discover sparse associations between concepts and the hidden-units of $f(\cdot)$. We express $k^{th}$ concept as a sparse vector $\rvv_{k} \in \mathbb{R}^L$ that represents a linear direction in the
intermediate space $\Phi_1(\cdot)$. 

(2) \emph{ Causal concept ranking}: Using tools from causal inference, we find an effective ranking of the concepts based on their relevance to the classification decision. Specifically, we consider each concept as a mediator in the causal path between the input and the outcome. We measure concept relevance as the effect of a counterfactual intervention on the outcome that passes indirectly through the concept-mediator.
       
  (3) \emph{Surrogate explanation function}: We learn an  easy-to-interpret function $g(\cdot)$  that mimics function $f(\cdot)$ in its decision. Using $g(\cdot)$, we seek to learn a global explanation for $f(\cdot)$ in terms of the concepts.

    \subsection{Concept associations}
    We discover concept associations with intermediate representation $\Phi_1(\cdot)$ by learning a binary classifier that maps $\Phi_1(\rvx)$ to the concept-labels~\cite{kim2018interpretability}. We treat each concept as a separate binary classification problem and extract a representative set of images $\mathcal{X}^{k}$, in which concept $c_n[k]$ is present and a random negative set. We define concept vector ($\rvv_{k}$) as the solution to the logistic regression model $ c_n[k] = \sigma(\rvv_k^T \vect(\Phi_1(\rvx_n))) + \epsilon$, where $\sigma(\cdot)$ is the sigmoid function. For a convolutional neural network, $\Phi_1(\rvx) \in \mathbb{R}^{w \times h \times l}$ is the output activation of a convolutional layer with width $w$, height $h$ and number of channels $l$. We experimented with two vectorization for $\Phi_1$. In first, we flatten $\Phi_1(\rvx)$ to be a $whl$-dimensional vector. In second, we applied a spatial aggregation by max-pooling along the width and height to obtain $l$-dimensional vector. \blue{Unlike TCAV~\cite{kim2018interpretability} that uses linear regression,} we  used lasso regression to enable sparse feature selection and minimize the following loss function,
 \begin{equation}\label{eq:matrixeqn2}
\min_{\rvv_k} \sum_{\rvx_n \in \mathcal{X}_k}  \ell(h_{\rvv_k}(\rvx), c_n[k]) + \lambda ||\rvv_{k}||_1
   \end{equation}    
    where $\ell(\cdot,\cdot)$ is the cross entropy loss, $h_{\rvv_k}(\rvx) = \sigma(\rvv_k^T \vect(\Phi_1(\rvx_n)))$ and $\lambda$ is the regularization parameter. 
We performed 10-fold nested-cross validation to find $\lambda$ with least error. The non-zero elements in the concept vector $\rvv_{k}$ forms the set of hidden units ($\mathcal{V}_{k}$) that are most relevant to the $k^{th}$ concept. 
%e call these hidden units as concept-units .

% repeated lasso regression multiple times with different samples from $\mathcal{X}_{c_k}$ and choose the final concept-units through majority voting.
%\[\mathcal{V}_{c_k} = \{m; v_{c_k}^{(m)} \neq 0 \}\]

 \subsection{Causal concept ranking}
% We quantify the relevance of a concept to the classification outcome, based measuring the causal effect of a concept on the outcome
 Concept associations identified hidden units that are strongly correlated with a concept. However, the neural network may or may not use the discovered concepts to arrive at its decision. We use tools from causal inference, to quantify what fraction of the outcome is mediated through the discovered concepts. 
 
 \blue{To enable causal inference, we first define  \textit{counterfactual} $\rvx'$ as a perturbation of the input image $\rvx$ such that the decision of the classifier is flipped. Following the approach proposed in \cite{singla2019explanation}, we used a conditional generative adversarial network (cGAN) to learn the counterfactual perturbation. We conditioned  on the output of the classifier, to ensure that cGAN learns a classifier-specific perturbation for the given image $\rvx$. Next, we used theory from causal mediation analysis to causally relate a concept with the classification outcome. Specifically, }%, while using the desired change in classifier outcome as the condition.} 
 %\textit{do}-operation on the input image. \textit{do}($\rvx$) denotes an intervention on input $\rvx$ to create a counterfactual 
 %such that prediction of the classifier flips as a result of the intervention. 
 %We obtain counterfactual for $\rvx$, using the counterfactual generation process proposed in \cite{singla2019explanation}. \blue{They trained a conditional generative adversarial network (cGAN) to learn a  perturbation of the input image such that the resulting image produces a desired outcome from the black-box classifier.} 
 we consider concept as a mediator in the causal pathway from the input $\rvx$ to the outcome $\rvy$. We specify following effects to quantify the causal effect of the counterfactual perturbation and the role of a mediator in transferring such effect, 
 
 \begin{enumerate}
     \item Average treatment effect (ATE): \blue{ATE  is the  total change in the classification outcome $\rvy$ as a result of the counterfactual perturbation.} 
     \item Direct effect (DE): DE is the effect of the \blue{counterfactual perturbation} that comprises of any causal mechanism that \textit{do not} pass through a given mediator. It captures how the \blue{perturbation of input image} changes classification decision directly, without considering a given concept.
      \item Indirect effect (IE): IE is the effect of the  \blue{counterfactual perturbation} which is mediated by a set of mediators. It captures how the  \blue{perturbation of input image} changes classification decision indirectly through a given concept.
 \end{enumerate}
 
 Following the potential outcome framework from \cite{rubin1974estimating,NEURIPS2020_92650b2e}, we define the ATE as the proportional difference between the factual and the counterfactual classification outcome,
  \begin{equation}\label{eq:ate}
\textbf{ATE} =  \mathbb{E}\big[\frac{f(\rvx')}{f(\rvx)}-1\big].
   \end{equation}   
 
\begin{figure}[htbp]
        \centerline{
        \includegraphics[width=0.9\textwidth]{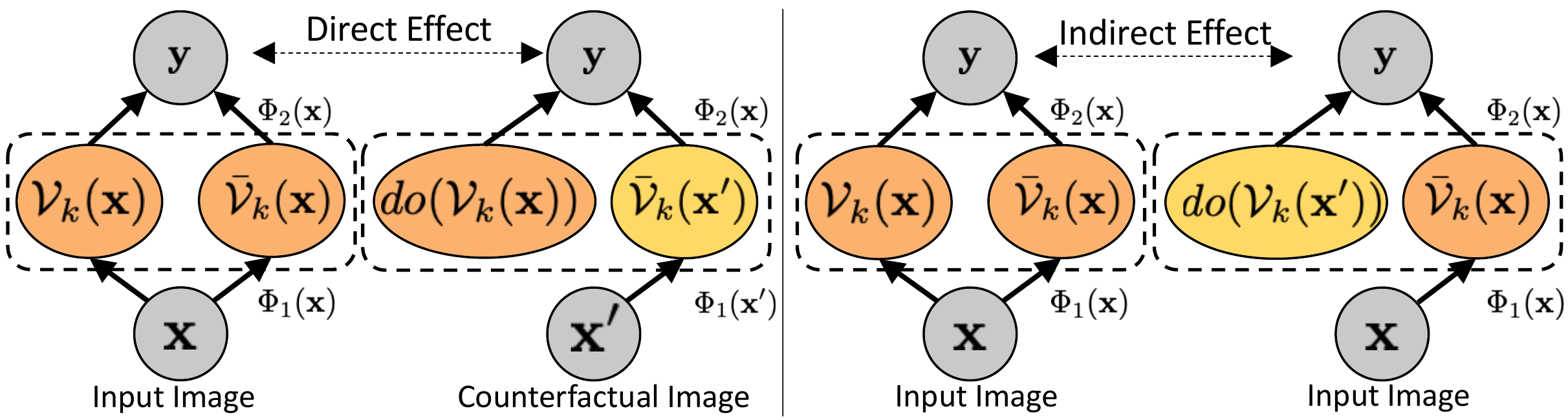}
        }
        \caption{Illustration of direct and indirect effects in causal mediation analysis.}
        \label{fig:cma1}
    \end{figure}

  To enable causal inference through a mediator, we  borrow Pearl’s definitions of natural direct and indirect effects~\cite{pearl2001direct} (\emph{ref} Fig.~\ref{fig:cma1}). We consider set of concept-units $\mathcal{V}_{k}$ as a mediator, representing the $k^{th}$ concept. We decompose the latent representation $\Phi_1(\rvx)$ as concatenation of response of concept-units \blue{$\mathcal{V}_k(\rvx)$} and rest of the hidden units \blue{$\Bar{\mathcal{V}}_k(\rvx)$ \ie   $\Phi_1(\rvx) = [\mathcal{V}_k \blue{(\rvx)} , \Bar{\mathcal{V}_k}(\rvx)]$}. \blue{We can re-write classification outcome as $f(\rvx) = \Phi_2(\Phi_1(\rvx)) = \Phi_2([\mathcal{V}_k(\rvx), \Bar{\mathcal{V}_k}(\rvx)])$}.
 \blue{To disentangle the direct effect from the indirect effect, we use the concept of \textit{do}-operation on the unit level of the learnt network. Specifically, we use 
   \textit{do}($\mathcal{V}_k\blue{(\rvx)}$) to denote that we set the value of the concept-units to the value obtained by using the original image as input. By intervening on the network and setting the value of the concept units, we can compute the direct effect as the proportional difference between the factual and the counterfactual classification outcome, while holding mediator \ie $\mathcal{V}_k$ fixed to its value before the perturbation,}

    \begin{equation}\label{eq:de}
    \blue{
\textbf{DE} =  \mathbb{E} \big[\frac{\Phi_2([do(\mathcal{V}_k(\rvx)), \Bar{\mathcal{V}}_k(\rvx')])}{\Phi_2([\mathcal{V}_k(\rvx) , \Bar{\mathcal{V}}_k(\rvx)])}-1\big].
}
   \end{equation} 
   
  We compute indirect effect as the expected change in the outcome, if we change  the mediator from its original value to its value using counterfactual, while holding everything else fixed to its original value,
  \begin{equation}\label{eq:ide}
  \blue{
\textbf{IE} = \mathbb{E} \big[\frac{\Phi_2([do(\mathcal{V}_k(\rvx')) , \Bar{\mathcal{V}}_k(\rvx)]) }{\Phi_2([\mathcal{V}_k(\rvx) , \Bar{\mathcal{V}}_k(\rvx)])}-1\big].
}
 \end{equation}
 
 %Specifically, we used causal mediation analysis (CMA)~\cite{robins1992identifiability,pearl2001direct} to measure the causal effect of a concept on an outcome. 
If the perturbation has no effect on the mediator, then the causal indirect effect will be zero. Finally, we use the indirect effect associated with a concept, as a measure of its relevance to the classification decision. 
 
 \subsection{Surrogate explanation function}
We aim to learn a surrogate function $g(\cdot)$, such that it reproduces the outcome of the function $f(\cdot)$  using an interpretable and straightforward function.
We formulated $g(\cdot)$ as a decision tree as many clinical decision-making procedures follow a rule-based pattern. We summarize the internal state of the function $f(\cdot)$ using output of $k$ concept regression functions $h_{\rvv_k}(\cdot)$ as follows, 
 \begin{equation}\label{eq:matrixeqn}
 \rvw_n = [,\text{logit}(h_{\rvv_1}(\rvx_n)) , \text{logit}(h_{\rvv_2}(\rvx_n)) , \cdots ].
   \end{equation}
Next, we fit a decision tree function, $g(\cdot)$, to mimic the outcome of the function $f(\cdot)$ as,
 \begin{equation}\label{eq:matrixeqn1}
 g^* = \arg\min_g \sum_n\mathcal{L}(g(\rvw_n), f(\rvx_n)),
   \end{equation}
where $\mathcal{L}$ is the splitting criterion based on minimizing entropy for  highest information gain from every split. 
 
% We further quantify the extend to which our explanation function $g(\rvw)$ approximates the decision of $f(\rvx)$. We adapt concept completeness score to quantify the part of the decision function $f(\cdot)$ that is expressible in-terms of concepts~\cite{yeh2019completeness} as, 
%\begin{equation}
 %   \eta_f(\rvv_{1}, \cdots, %\rvv_{k}) = \frac{ %\mathbb{P}_{\rvx, \rvy \sim %\mathcal{X}_{test}}[y = %\arg \max_{y'}g(\rvw)]-\alp%ha_r}{\mathbb{P}_{\rvx, %\rvy \sim %\mathcal{X}_{test}}[y = %\arg \max_{y'}f(\rvx)]-\alp%ha_r},
%\end{equation}
%where $\mathbb{P}_{\rvx, \rvy \sim \mathcal{X}_{test}}[y = \arg \max_{y'}g(\rvw)]$ is the accuracy achieved by the surrogate model $g(\cdot)$ and $\alpha_r$ is the accuracy of random prediction, which forms the lower bound of the completeness score to 0. 
\section{Experiments}
We first evaluated the concept classification performance and visualized  concept-units to demonstrate their effectiveness in localizing a concept.   Next, we summarized the indirect effects associated with different concepts across different layers of the classifier. We evaluated a proposing ranking of the concepts based on their causal contribution to the classification decision. Finally, we used  the top-ranked concepts to learn a surrogate explanation function in the form of a decision tree.
 \begin{figure}[t]
        \centerline{
        \includegraphics[width=0.9\textwidth]{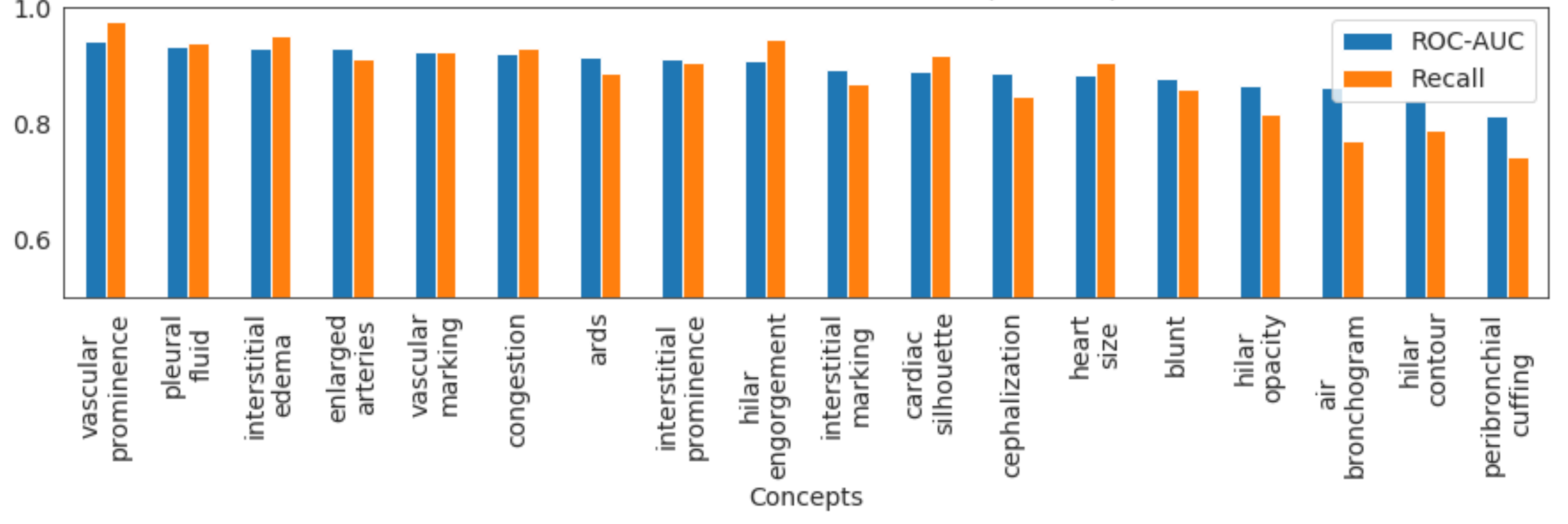}
        }
        \caption{AUC-ROC and recall metric for different concept classifiers.}
        \label{fig:lasso}
    \end{figure}
\textit{Data preprocessing:} We perform experiments on the MIMIC-CXR~\cite{johnson2019mimic} dataset, which is a multi-modal dataset consisting of 473K chest X-ray images and 206K reports. The dataset is labeled for 14 radiographic observations, including 12 pathologies. We used state-of-the-art DenseNet-121~\cite{huang2017densely} architecture for our classification function~\cite{irvin2019chexpert}. DenseNet-121 architecture is composed of four dense blocks. We experimented with three versions of $\Phi_1(\cdot)$ to represent the network until the second, third, and fourth dense block. For concept annotations, we considered radiographic features that are frequently mentioned in radiology reports in the context of labeled pathologies. Next, we used Stanford CheXpert~\cite{irvin2019chexpert} to extract and classify these observations from free-text radiology reports. 
%We prune the concept list to remove observations with fewer than 300 positive mentions. 
%We applied our algorithm on $K = 18$ concepts for three diagnosis namely, cardiomegaly, pleural effusion and edema.
   \begin{figure}[h]
        \centerline{
        \includegraphics[width=0.9\textwidth]{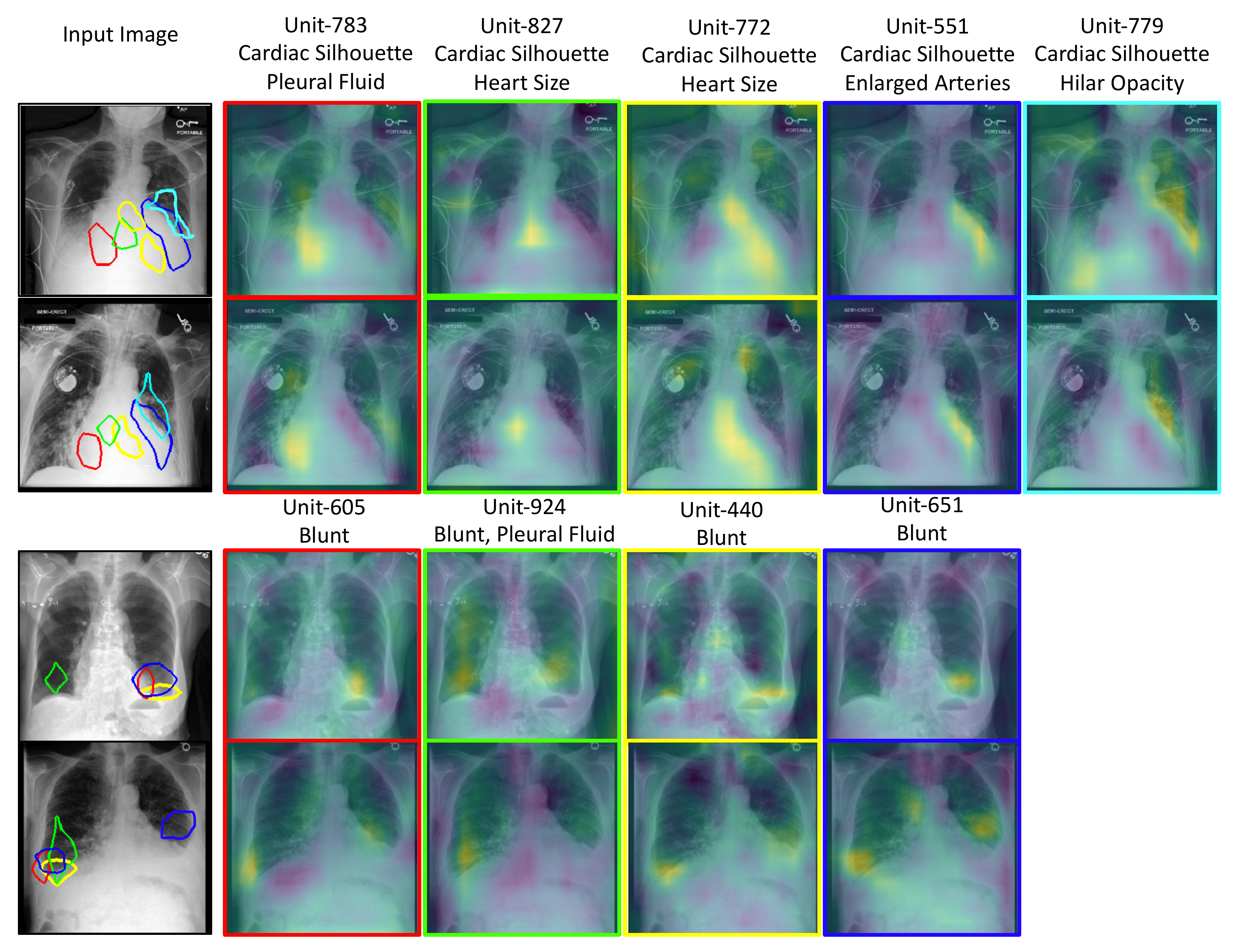}
        }
        \caption{A qualitative demonstration of the activation maps of the hidden units that act as visual concept detectors. \blue{Each column represents one hidden unit identified as part of concept vector $\mathcal{V}_k$. Top two rows show $k=$ \textit{cardiac-silhouette} and bottom rows have $k=$\textit{blunt costophrenic angle}}. }
        \label{fig:as}
    \end{figure} 
\subsection{Evaluation of concept classifiers}

The intermediate representations from third dense-block consistently outperformed other layers in concept classification. In Fig.~\ref{fig:lasso}, we show the testing-ROC-AUC and recall metric for different concept classifiers. All the concept classifiers achieved high recall, demonstrating a low false-negative (type-2) error.

 \blue{In Fig.~\ref{fig:as},  we visualize the activation map of hidden units associated with the concept vector $\mathcal{V}_k$. For each concept, we visualize hidden units that have large logistic regression-coefficient ($\beta_k$). To highlight the most activated region for a unit, we threshold activation map by the top 1\% quantile of the distribution of the selected units' activations~\cite{bau2017network}. Consistent with prior work \cite{bau2020understanding}, we observed that several hidden units have emerged as concept detectors, even though concept labels were not used while training $f$. For \textit{cardiac-silhouette}, different hidden units highlight different regions of the heart and its boundary with the lung. For localized concept such as \textit{blunt costophrenic angle}, multiple relevant units were identified that all focused on the lower-lobe regions. Same hidden unit can be relevant for multiple concepts. The top label in Fig.~\ref{fig:as}. shows the top two important concepts for each hidden unit.}

 %associated with concepts.
 
 %for a given concept, we visualize the activation map of hidden units with logistic regression-coefficient ($\beta_k$).

 %For a given concept, we visualize top hidden units sorted based on logistic regression-coefficient ($\beta_k$).
 
 %that have with large coefficient learned in logistic regression. We show the top-activated images for the selected hidden unit. 
 \blue{} 
 \begin{figure}[htbp]
        \centerline{
        \includegraphics[width=1.0\textwidth]{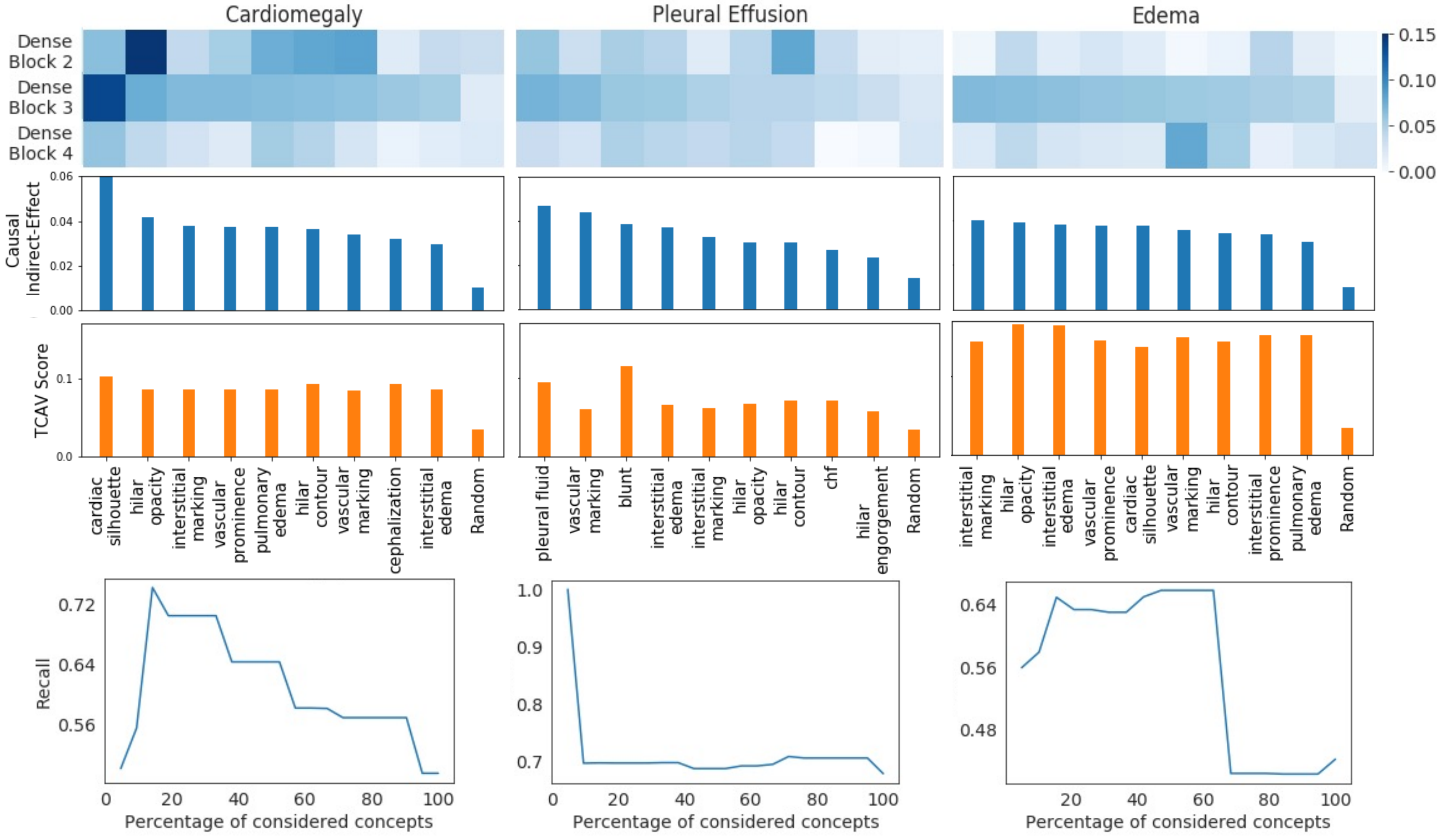}
        }
        \caption{Indirect effects of the concepts, calculated over different layers of the DenseNet-121 architecture (heat-map). The derived  ranking of the concepts based on their causal relevance to the diagnosis (bar-graph). A comparative ranking based on concept sensitivity score from TCAV~\cite{kim2018interpretability}. The trend of recall metric for the decision tree function $g(\cdot)$, while training using top x\% of top-ranked concepts (trend-plot).}
        \label{fig:cma}
    \end{figure}

\subsection{Evaluating causal concepts using explanation function}

%In-direct effect captures the strength of the causal relationship between a mediator (concept) and the outcome (class-label). 
We evaluate the success of the counterfactual intervention by measuring ATE. High values for ATE confirms that counterfactual image generated by~\cite{singla2019explanation} successfully flips the classification decision. We achieved an ATE of 0.97 for cardiomegaly, 0.89 for pleural effusion and 0.96 for edema. In Fig.~\ref{fig:cma} (heat-map), we show the distribution of the indirect effect associated with concepts, across different layers. The middle layer demonstrates a large indirect effect across all concepts. This shows that the hidden units in dense-block 3 played a significant role in mediating the effect of counterfactual intervention.

 In Fig.~\ref{fig:cma} (bar-graph), we rank the concepts based on their indirect effect. The top-ranked concepts recovered by our ranking are consistent with the radiographic features that clinicians associates with the examined three diagnoses~\cite{karkhanis2012pleural,milne1985radiologic,nakamori1991effect}. \blue{Further, we used the concept sensitivity score from TCAV~\cite{kim2018interpretability} to rank concepts for each diagnosis. The top-10 concepts identified by our in-direct effect and TCAV are the same, while their order is different. The top-3 concepts are also the same, with minor differences in ranking. Both the methods have low importance score for random concept. This confirms that the trend in importance score is unlikely to be caused by chance. For our approach, random concept represents an ablation of the concept-association step. Here, rather than performing lasso regression to identify relevant units, we randomly select units. }

\begin{figure}[htbp]
\centerline{
        \includegraphics[width=0.75\textwidth]{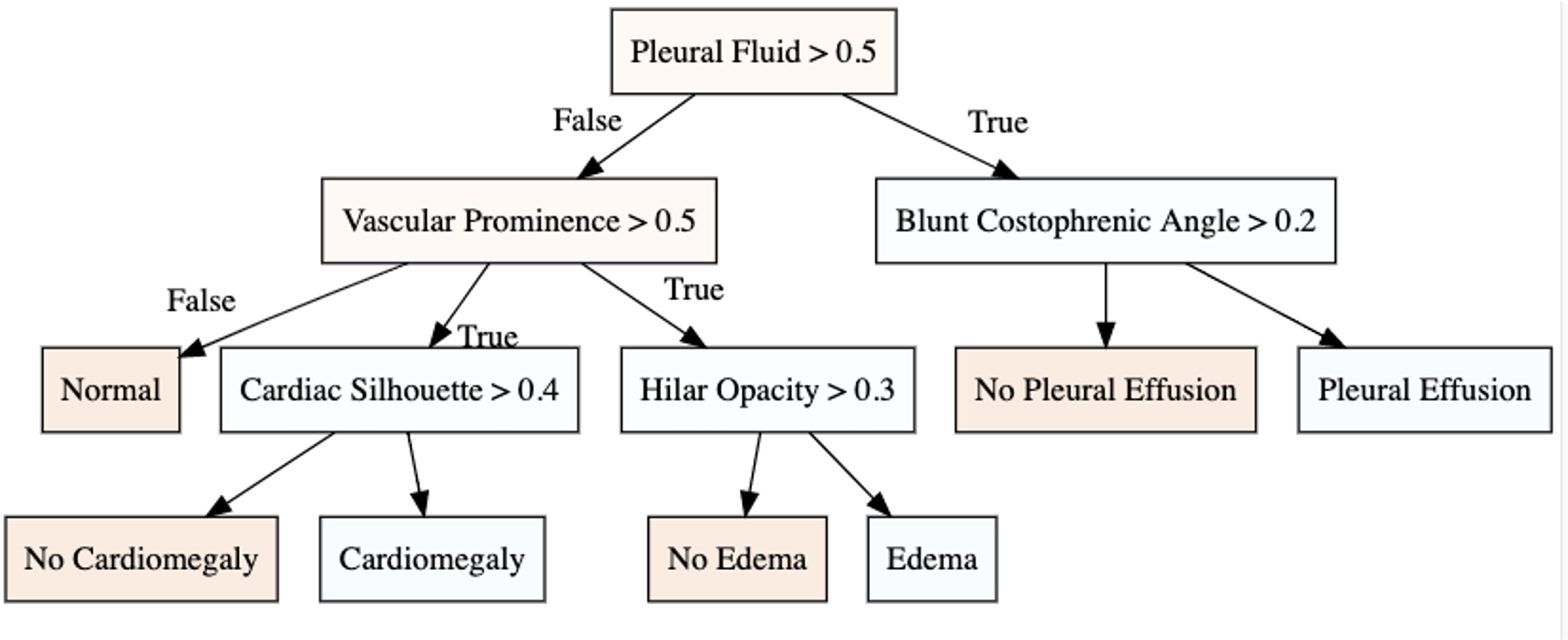}
        }
        \caption{The decision tree for the three diagnosis with best performance on recall metric.}
        \label{fig:dt1}
\end{figure}

 To quantitatively demonstrate the effectiveness of our ranking, we iteratively consider $x\%$ of top-ranked concepts and retrain the explanation function $g(\rvw)$. In Fig.~\ref{fig:cma} (bottom-plot), we observe the change in recall metric for the classifier $g(\cdot)$ as we consider more concepts. In the beginning, as we add relevant concepts, the true positive rate increases resulting in a high recall. However, as less relevant concepts are considered, the noise in input features increased, resulting in a lower recall. Fig.~\ref{fig:dt1} visualize the decision tree learned for the best performing model. %Finally, we summarize our explanation framework in Fig.~\ref{fig:dt_ex}.  
 
 %We achieved a concept completeness score ($\eta_f$) of 0.17 for Cardiomegaly, 0.25 for Pleural effusion and 0.26 for Edema. In future work, we would consider more concepts to improve $\eta_f$.

    % The decision tree provides global explanation for a class label. $\rvv_c$ provides local explanation for a sample $\rvx$ in terms of important concepts for current sample.

\section{Conclusion}
We proposed a novel framework to derive global explanation for a black-box model. Our explanation is grounded in terms of clinically relevant concepts that are causally influencing the model's decision. As a future direction, we plan to extend our definition of concepts to include a broader set of clinical metrics.

\textit{Acknowledgement} This work was partially supported by NIH Award Number 1R01HL141813-01, NSF 1839332 Tripod+X, SAP SE, and Pennsylvania's Department of Health. We are grateful for the computational resources provided by Pittsburgh SuperComputing grant number TG-ASC170024.

%\bibliography{main}

%\bibliographystyle{splncs04}

 \bibliographystyle{splncs04}
 \bibliography{main}

\end{document}